\documentclass[letterpaper]{article} 
\usepackage{aaai2027}  
\usepackage[hyphens]{url}  
\usepackage{graphicx} 
\usepackage{natbib}  
\usepackage{caption} 
\usepackage{amsmath}
\usepackage{amssymb}
\usepackage{booktabs}
\usepackage{tabularx}

\newcommand{\latedit}{\ensuremath{\langle\mathrm{LAT\_EDIT}\rangle}}
\newcommand{\latprop}{\ensuremath{\langle\mathrm{LAT\_PROP}\rangle}}
\newcommand{\lateffect}{\ensuremath{\langle\mathrm{LAT\_EFFECT}\rangle}}
\title{Localize, Then Reason: Visual Latent Structural Reasoning for Molecular Properties and Edits}
\author{
    Xingqiao Lin\textsuperscript{\rm 1,\rm 2},
    Junmei Wang\textsuperscript{\rm 2}\corresponding,
    Haocheng Tang\textsuperscript{\rm 3}\corresponding
}
\affiliations{
    \textsuperscript{\rm 1}Department of Chemical Engineering, Carnegie Mellon University, Pittsburgh, Pennsylvania 15213, United States\\
    \textsuperscript{\rm 2}Department of Pharmaceutical Sciences and Computational Chemical Genomics Screening Center, School of Pharmacy, University of Pittsburgh, Pittsburgh, PA, USA\\
    \textsuperscript{\rm 3}Khoury College of Computer Science, Northeastern University, Boston, MA, USA
}

\begin{document}

\maketitle

\begin{abstract}

Local chemical perception and property reasoning are both essential for understanding how molecular structure determines properties. Current LLM-based chemical reasoning methods either receive SMILES/molecular images together with descriptions of local motifs, or reason directly from molecular images. Neither approach enables the model to focus on chemically meaningful regions before reasoning. To address this gap, we propose Visual Latent Structural Reasoning (VLSR), an end-to-end framework that jointly learns localization and reasoning from molecular images. Central to our approach is a localize-then-reason strategy. VLSR first learns to locate chemically meaningful regions in a molecular image. It then reasons about their property effects in a compact latent workspace before producing the final answer. Under the same inference setup, this design achieves $9.6\times$ higher throughput than a comparable textual-reasoning baseline.

\end{abstract}

\section{Introduction}
\label{sec:introduction}

Local structure perception is fundamental to molecular property prediction. 
Recent approaches increasingly incorporate molecular structural information into LLM-based models by providing multi-modal representations, such as SMILES with local motif annotations. 
However, these annotations explicitly reveal the relevant structural components to the model, reducing the need for the model to discover which regions of a molecule are chemically responsible for the queried property. 
This motivates a different paradigm: instead of providing structural explanations as inputs, a model should learn to localize chemically meaningful regions and subsequently reason over their property effects.

Recent chemical LLMs and vision-language models (VLMs) can reason over increasingly rich molecular inputs \citep{zhang2024chemllm,li2024chemvlm,tan2025chemmllm}. A common approach is to input molecule structural descriptors. MSR provides extracted structural components, such as functional groups, as natural-language descriptions \citep{jang2025msr}. ChemVLR incorporates functional-group information into the textual Chain-of-Thought (CoT) used for training \citep{zhao2026chemvlr}. MPPReasoner jointly processes molecular images and SMILES and rewards textual reasoning that describes tool-identified local structures \citep{zhuang2025mppreasoner}. These methods teach the model to describe local chemistry in language. However, training a model to describe a functional group is not equivalent to learning how local structure influences the queried property.


\begin{figure}[t]
    \centering
    \includegraphics[width=1.\columnwidth]{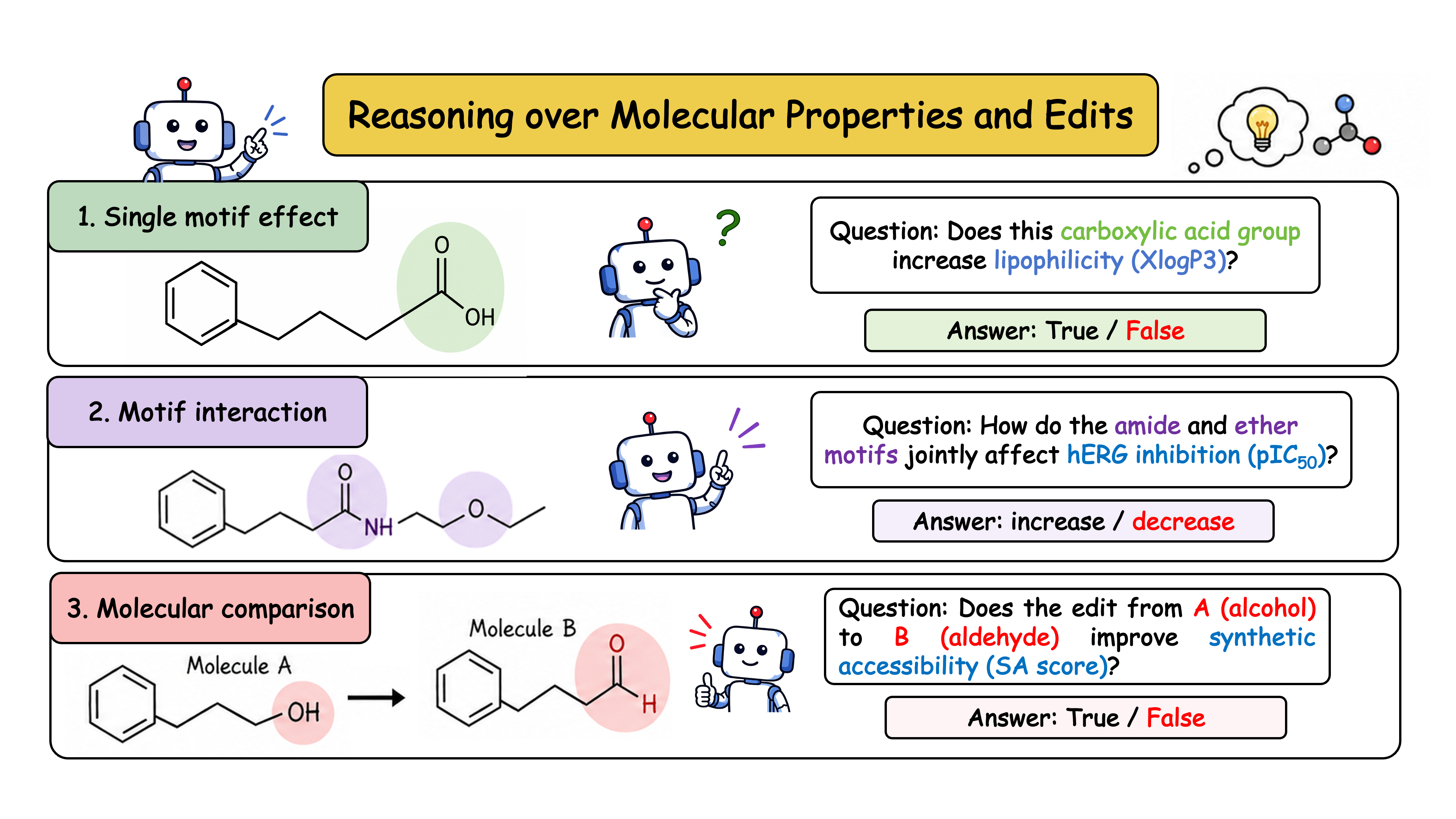}
    \caption{Three molecular property reasoning tasks: single-motif effects, motif interactions, and molecular comparisons.}
    \label{fig:task}
\end{figure}

Visual localization provides a natural solution for this gap. 
In optical chemical structure recognition (OCSR), MolScribe localizes atoms before reconstructing molecular graphs, while GTR-VL models graph traversal through visual Chain-of-Thought reasoning \citep{qian2023molscribe,wang2025gtrcot}. 
These works demonstrate that molecular depictions contain recoverable localized evidence. 
However, their objective is structural reconstruction: they recover what is present in an image, but not which regions matter for a downstream chemical question.

This motivates the principle of \emph{localize, then reason}. 
Rather than providing predefined motifs as explicit inputs, VLSR learns to identify property-relevant molecular regions as intermediate representations for downstream reasoning. 
The key challenge is not merely decomposing a molecule into structural components, but discovering which localized chemical patterns are informative for a specific property query and how they collectively contribute to the prediction. 
Accordingly, VLSR learns region-level representations under property supervision, where localization serves as an adaptive interface between molecular perception and property reasoning.

Substructure annotations supervise region localization during training, but their boxes and labels are not included in the reasoning path or provided at inference. The region-level representations are processed together with the property question in a structured latent workspace. After multiple workspace updates, the model decodes only the final answer rather than an intermediate textual rationale.

\begin{figure*}[t]
    \centering
    \includegraphics[width=\textwidth]
    {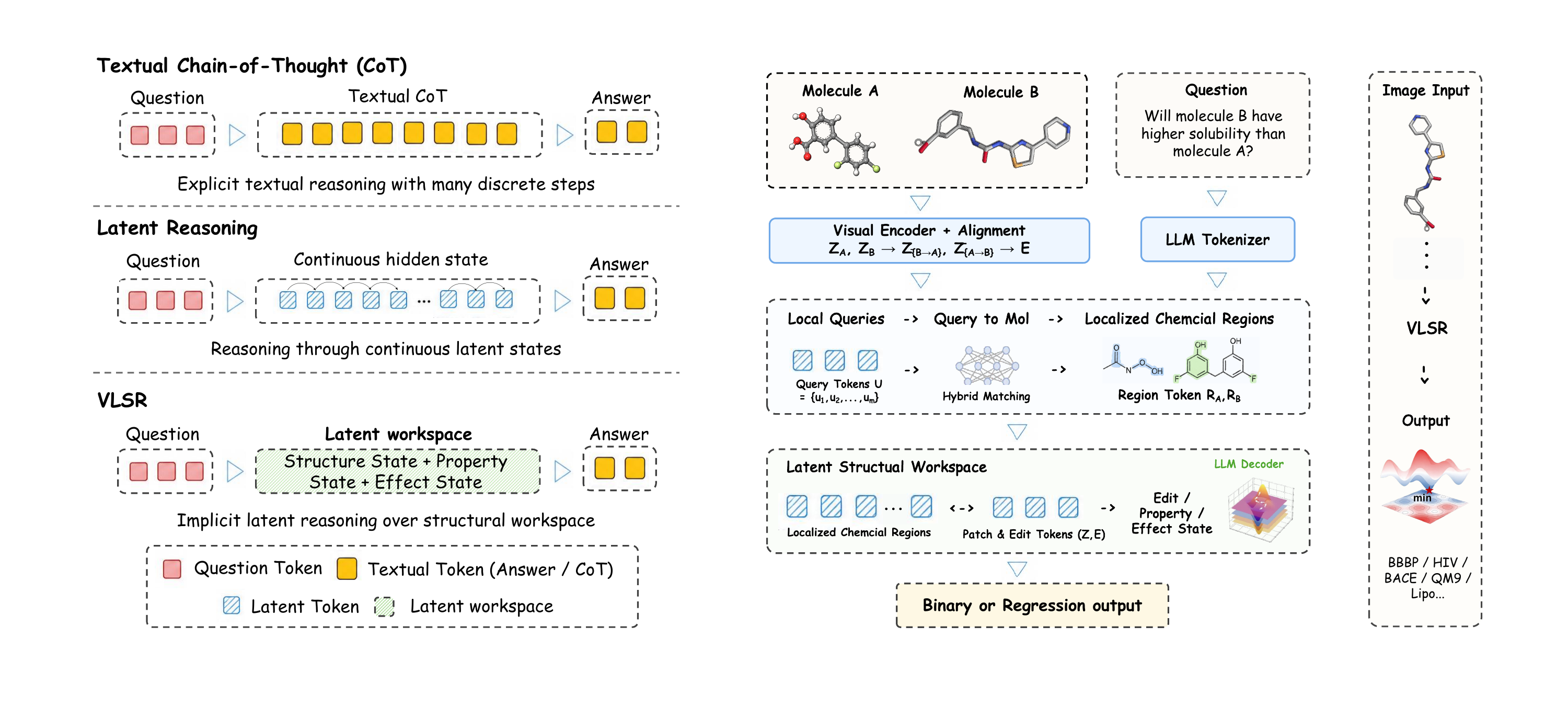}
    \caption{Left: Comparison of textual Chain-of-Thought, continuous latent reasoning, and VLSR. Right: Detailed illustration of the VLSR framework.}
    \label{fig:latent}
\end{figure*}

Our contributions are threefold:
\begin{itemize}
    \item We introduce a \emph{localize-then-reason} formulation for molecular property reasoning, in which the model learns to find local chemical structures before reasoning about their property effects.

    \item We develop VLSR, which combines learned chemical localization with a compact latent workspace that reasons without autoregressively generating intermediate text.
    
    \item VLSR outperforms specialized chemical models and substantially larger general-purpose LLMs. Without additional training, it can generalize to Schrödinger's multiple FEP binding-affinity benchmarks.
\end{itemize}

\section{Related Work}
\label{sec:related-work}

\subsection{Molecular Structure Reasoning with LLMs}

Chemical LLMs incorporate explicit structure either as model input or as CoT supervision. LLM-MPP and LLaMo combine SMILES, molecular graphs, and language, making graph structure directly available to the LLM \citep{jin2025llmmpp,park2024llamo}. MSR provides extracted components such as functional groups as natural-language descriptions, while TreeKD converts functional-group features into textual rules for property prediction \citep{jang2025msr,le2026treekd}. ChemVLR incorporates functional-group information into the textual CoT used for training \citep{zhao2026chemvlr}. Similarly, MPPReasoner inputs a molecular image together with SMILES and rewards reasoning that describes tool-identified functional groups \citep{zhuang2025mppreasoner}. These approaches teach the model to express local chemistry through graphs or language. VLSR instead follows a \emph{localize-then-reason} strategy: the model learns to locate relevant local structures in the molecular image and then determine how they affect the queried property. 

\subsection{Molecular Images and Local Structure}

ChemVLM and ChemMLLM adapt multimodal language models to molecular images and chemical tasks, while ChemDFM-X aligns multiple chemical modalities \citep{li2024chemvlm,tan2025chemmllm,zhao2024chemdfmx}. TinyChemVL further improves efficiency by reducing visual tokens \citep{zhao2025tinychemvl}. ChemSeek-OCR attempts to directly transfer the OCR model DeepSeek-OCR-2 to the OCSR task \citep{tang2026finetuning}. These models demonstrate that molecular depictions can support chemical reasoning, but their objectives do not explicitly require the model to locate local chemical structures before predicting a property. Chemical localization has been studied for other purposes. SubGrapher segments functional groups and carbon backbones for molecular-image retrieval \citep{morin2025subgrapher}, while OCSR methods locate atoms and bonds to reconstruct molecular graphs or structural strings \citep{qian2023molscribe,morin2023molgrapher,rajan2023decimer}. VLSR connects localization with property reasoning: the model first learns to find local chemical structures and then reasons about their effects on the queried property.

Post hoc graph explainers provide a complementary notion of chemical
localization. GNNExplainer identifies compact subgraphs and node features that
preserve a graph model's prediction, whereas SubgraphX explicitly searches for
important subgraphs and accounts for interactions among them
\citep{ying2019gnnexplainer,yuan2021subgraphx}. These approaches explain a
prediction over an explicit graph after it has been produced. VLSR instead
learns regions directly from depiction pixels and uses them as intermediate
computational states before the property prediction. Consequently, spatial
agreement alone is not sufficient evidence of useful reasoning; the relational
controls and occlusion intervention test whether the localized regions also
support the downstream decision. The regions are therefore part of the model's
computation, not merely a visualization of its output.

\subsection{Textual and Latent Chemical Reasoning}

CoT expresses intermediate reasoning through generated text \citep{wei2022cot}. Although text can name functional groups and explain chemical effects, it may not precisely preserve fine-grained structural details, such as attachment sites, neighboring context, and correspondences between molecular changes. Continuous latent reasoning offers an alternative by maintaining intermediate information in hidden states. Coconut introduces continuous reasoning for general tasks, while LatentChem applies latent computation to chemical problems \citep{hao2025coconut,ye2026latentchem}. VLSR introduces a structured latent workspace for molecular property reasoning, with distinct latent roles for structural evidence, the queried property, and the predicted effect. This organization allows the model to connect visual structure with property outcomes without converting every intermediate detail into text.

\section{Problem Formulation}
\label{sec:problem-formulation}

We formulate molecular property reasoning from depictions and a property query. Each example contains a natural-language query $q$ and either one molecular image $I_A$ or a pair $(I_A,I_B)$. Single-molecule questions ask how local structure affects a property; paired questions ask how an edit changes that property. The answer $y$ may be a binary or directional label, a categorical answer, or a continuous value:
\begin{equation}
\hat{y}=f_\theta(I_A,I_B,q),
\end{equation}
where $I_B$ is omitted for single-molecule tasks. We require the model to derive local structure $R$ from the images before estimating $p_\theta(y\mid I_A,I_B,q,R)$. RDKit-derived boxes and labels \citep{landrum2026rdkit} are training targets rather than observed variables in this conditional distribution. They include functional groups, ring systems, and hetero-atom regions. At test time, the model must locate these structures, preserve their context, and identify edits directly from the depictions. Localization guides the prediction but does not remove access to the full image.

\section{Method}
\label{sec:method}

\subsection{Overview}
\label{sec:overall-framework}

VLSR receives one or two molecular depictions and a property query. It
encodes image patches, aggregates them into candidate chemical regions,
and, for paired inputs, aligns the two depictions before interpreting
their difference. Three latent tokens---\latedit, \latprop, and
\lateffect---then organize structural evidence, the queried property,
and its predicted effect. The pipeline is
\begin{equation}
(I,q)\ \mathrm{or}\ (I_A,I_B,q)
\rightarrow Z
\rightarrow (R,E)
\rightarrow
L
\rightarrow
y,
\end{equation}
where $Z$, $R$, $E$, and $L$ denote image tokens, localized regions,
edit evidence, and latent reasoning states, respectively; $E$ is omitted
for single-image inputs.

\subsection{Generic Patch Encoding}
\label{sec:visual-molecular-perception}

The Qwen-VL vision encoder \citep{bai2023qwenvl,qwen2026qwen35} maps each
depiction to patch tokens:
\begin{equation}
Z_A=f_{\mathrm{vision}}(I_A), \qquad
Z_B=f_{\mathrm{vision}}(I_B),
\end{equation}
where $Z_A,Z_B\in\mathbb{R}^{N\times d}$; single-molecule tasks use only
$Z_A$. Because patch boundaries need not coincide with chemical
structures, VLSR next forms region-level representations.

\subsection{Chemical Region Localization}
\label{sec:image-grounded-region}

Given $Z\in\mathbb{R}^{N\times d}$, trainable region queries $U$ retrieve
local chemical evidence through cross-attention:
\begin{equation}
A=\operatorname{softmax}\!\left(
\frac{(UW_Q)(ZW_K)^\top}{\sqrt d}
\right),
\qquad
R=AZW_V.
\end{equation}

The attention maps $A$ provide soft spatial grounding, while $R$ contains
region tokens that can integrate multi-patch motifs and ring context.
To obtain localization supervision, we derive chemical regions from RDKit
molecular graphs, including functional groups, ring systems, and
hetero-atom-centered regions. Their atom coordinates are projected onto
the 2D molecular depiction to obtain image-space bounding boxes and region
types. 

RDKit-derived regions supervise this bottleneck during training. A
Hungarian one-to-one assignment \citep{kuhn1955hungarian} provides the
primary matching, supplemented by a lightweight one-to-many hybrid branch
\citep{jia2023hdetr}. Boxes and region labels are auxiliary targets only:
they are not passed to the reasoning core and are absent at inference.
\subsection{Correspondence Before Edit Reasoning}
\label{sec:cross-image-edit}

For paired questions, VLSR uses bidirectional soft alignment rather than
index-wise patch subtraction. For $(X,Y)\in\{(A,B),(B,A)\}$,

\begin{equation}
\begin{aligned}
&\bar Z_{Y\rightarrow X}
=\mathrm{CrossAttn}(Z_X,Z_Y,Z_Y),\\
&\Delta_X=\bar Z_{Y\rightarrow X}-Z_X,\qquad
P_X=\bar Z_{Y\rightarrow X}\odot Z_X,\\
&E_X=\mathrm{MLP}_{\mathrm{edit}}
\big[Z_X,\bar Z_{Y\rightarrow X},\Delta_X,P_X\big].
\end{aligned}
\end{equation}
Learned pooling yields $E=\mathrm{EditPool}([E_A;E_B])$. The difference
emphasizes change, the product retains shared context, and the two
directions preserve additions and removals without requiring atom maps.

\begin{figure*}[t]
    \centering
    \includegraphics[width=\textwidth]{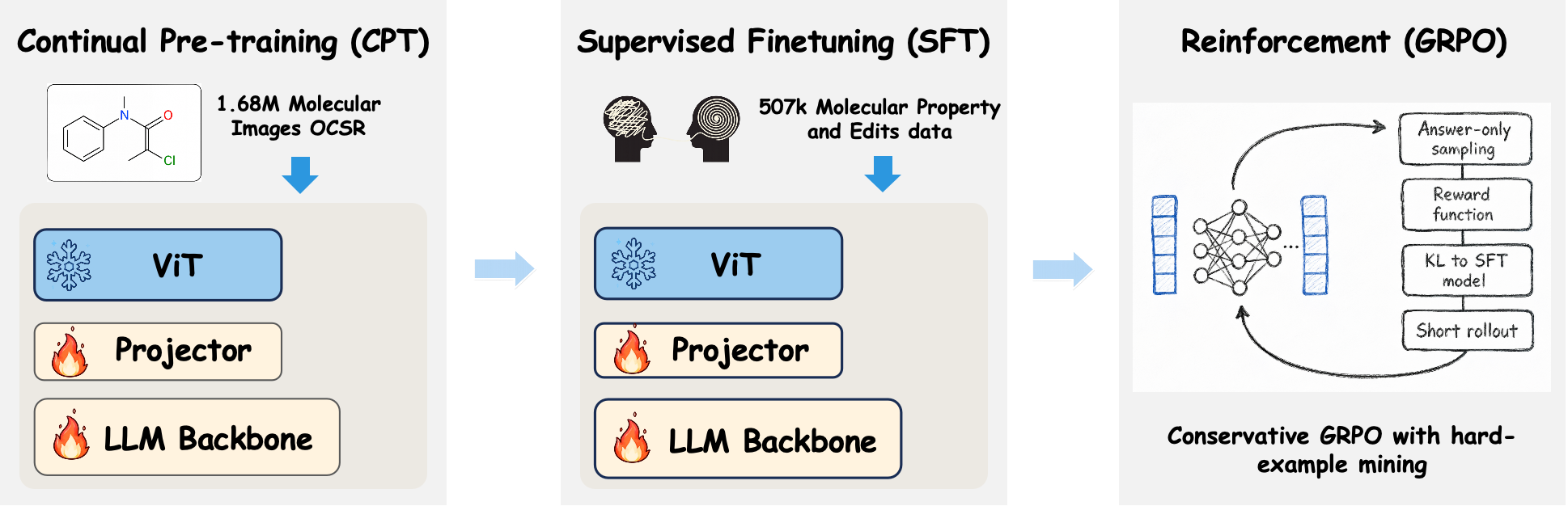}
    \caption{Training pipeline of VLSR. We first perform OCSR-style continued pre-training on large-scale molecular image data, then conduct supervised fine-tuning, and finally apply GRPO with answer-level rewards.}
    \label{fig:training-pipeline}
\end{figure*}

\subsection{Reasoning over Local Evidence}
\label{sec:visual-latent-structural-reasoning-workspace}

After the language backbone encodes the prompt, the hidden states at the
three special-token positions initialize the latent workspace:
\begin{equation}
(L_{\mathrm{edit}}^0,L_{\mathrm{prop}}^0,L_{\mathrm{effect}}^0)
=
\operatorname{Gather}(H_q;p_{\mathrm{edit}},p_{\mathrm{prop}},p_{\mathrm{effect}}),
\end{equation}
where $H_q$ contains the prompt states. With the question states $Q$,
image tokens $Z$, region tokens $R$, and optional edit tokens $E$, the
reasoning core performs $M$ Transformer updates:
\begin{equation}
\begin{aligned}
H^0&=[Q,Z,R,E,
L_{\mathrm{edit}}^0,L_{\mathrm{prop}}^0,
L_{\mathrm{effect}}^0],\\
H^{\ell+1}&=\operatorname{TransformerLayer}_{\ell}(H^{\ell}),
\quad \ell<M.
\end{aligned}
\end{equation}
The final edit, property, and effect states condition answer generation.
They preserve the roles of structural evidence, query semantics, and
predicted outcome without decoding an intermediate textual rationale.

\subsection{Training Strategy}
\label{sec:training-strategy}

VLSR uses three training stages.

\paragraph{Stage I: Continued Pre-training.}

We perform OCSR-style continued pre-training on one million synthetic
PubChem molecules \citep{kim2023pubchem} and 680K patent examples,
following MolScribe-style data construction \citep{qian2023molscribe}.
The vision encoder remains frozen.

\paragraph{Stage II: Supervised Fine-tuning.}

We construct FGBench-Scaffold from FGBench \citep{liu2025fgbench}
using pair-aware Bemis--Murcko scaffold components
\citep{wu2018moleculenet,bemis1996frameworks}, an ECFP4 similarity
threshold of 0.7, and stratified component allocation. Its 625,936
examples comprise 507,000 training, 56,348 validation, and 62,588 test
instances. LoRA \citep{hu2022lora} jointly optimizes answer prediction
and auxiliary region localization.

To evaluate molecular generalization under reduced scaffold leakage,
we split molecular pairs into indivisible chemical components before
train/validation/test allocation. The resulting split prevents
component overlap between training and test sets and enforces a maximum
ECFP4 similarity of 0.7 between test and training components. Compared
with the original FGBench partition, this reduces the nearest-training
similarity from 0.8596 to 0.4127 while eliminating test units with
similarity above 0.7.

\paragraph{Stage III: GRPO Optimization.}

We apply Group Relative Policy Optimization (GRPO)
\citep{shao2024deepseekmath} to a deterministically sampled subset of SFT
training records, stratified by answer outcome, property class, and action
type. The answer-level reward combines correctness, regression closeness, sign
consistency, and output cleanliness; it does not access annotations,
attention maps, or latent states.

\subsection{Learning Objective}
\label{sec:learning-objective}

The answer loss is
\begin{equation}
\mathcal{L}_{\mathrm{answer}}
=
-\sum_{t=1}^{|y|}
\log p_\theta(y_t \mid y_{<t},x).
\end{equation}
The supervised objective is
\begin{equation}
\begin{aligned}
\mathcal{L}_{\mathrm{total}}
={}&\mathcal{L}_{\mathrm{answer}}
+0.05\mathcal{L}_{\mathrm{attn}}^{\mathrm{ce}}
+0.05\mathcal{L}_{\mathrm{attn}}^{\mathrm{iou}}\\
&+0.01\mathcal{L}_{\mathrm{div}}
+0.02\mathcal{L}_{\mathrm{ent}}
+0.02\mathcal{L}_{\mathrm{fg}}\\
&+0.02\mathcal{L}_{\mathrm{hyb}}^{\mathrm{ce}}
+0.02\mathcal{L}_{\mathrm{hyb}}^{\mathrm{iou}}
+0.01\mathcal{L}_{\mathrm{hyb}}^{\mathrm{fg}} .
\end{aligned}
\end{equation}
For Hungarian-matched pairs, attention cross-entropy and soft-IoU
supervise coverage, diversity and entropy regularize the maps, and
$\mathcal{L}_{\mathrm{fg}}$ predicts the region type. The hybrid branch
applies the corresponding attention and type losses to its one-to-many
matches. Losses are averaged over valid pairs. All targets and auxiliary
heads are removed at inference.

\section{Experiments}
\label{sec:experiments}

The evaluation tests the localize-then-reason claim. We first compare learned localization with symbolic inputs and generic image patches. We then use random-region and component ablations to ask whether the model depends on meaningful local evidence and whether it can use that evidence for reasoning. Renderer shifts test whether localization survives changes in depiction layout, while an independent protein-conditioned benchmark tests transfer beyond the training task.

\subsection{Experimental Setup}

We evaluate all methods on FGBench-Scaffold, our pair-aware
scaffold-component split of FGBench. The benchmark contains questions about
single-motif effects, motif interactions, and molecular comparisons. Each
example provides a property query together with one or two molecular
representations. Image-input models receive RDKit-rendered depictions, whereas
symbolic baselines receive the corresponding SMILES. The required answer is
either a binary decision or a continuous property change.

\subsection{Baselines}

We compare VLSR with both SMILES-input and image-input baselines. The symbolic group includes instruction-based chemical LLMs, LatentChem \citep{ye2026latentchem}, and Qwen3.5-4B models \citep{qwen2026qwen35} fine-tuned on the same reasoning data. The visual group includes general VLM backbones, ChemVLM-26B \citep{li2024chemvlm}, ChemDFM-X \citep{zhao2024chemdfmx}, and supervised Qwen3.5-4B variants. The protocols cover direct prediction, explicit Think-style textual reasoning, and latent reasoning. We also evaluate Molecular Structural Reasoning (MSR), which supplies an LLM with an extracted textual description of molecular structure \citep{jang2025msr}. All task-specific baselines are trained and evaluated using the same FGBench-Scaffold split.

\subsection{Main Results}

\begin{figure*}[t]
    \centering
    \includegraphics[width=\textwidth]{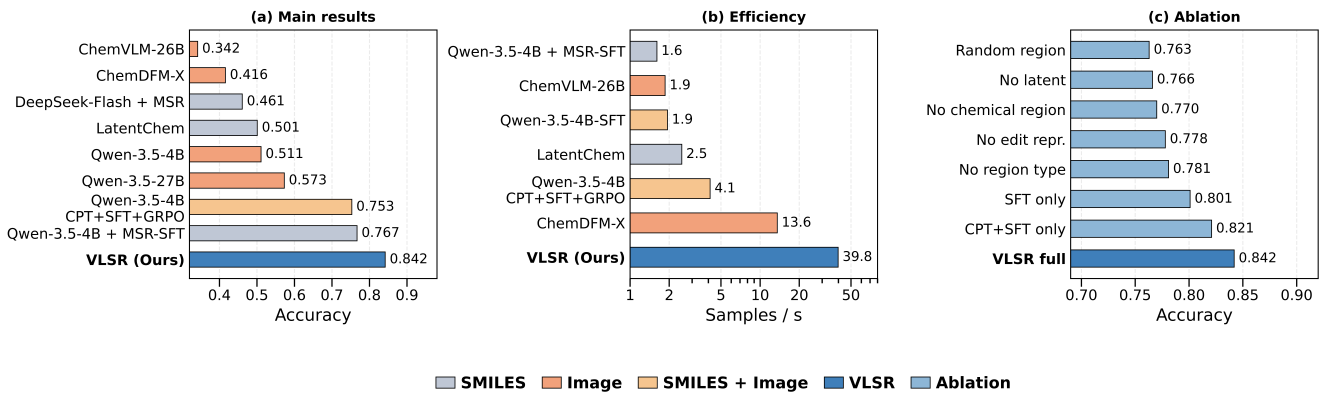}
    \caption{Summary of experimental results. (a) Accuracy comparison with representative SMILES-input, image-input, and chemical VLM baselines. (b) Inference throughput on the full test set. (c) Accuracy drop under ablations, where larger drops indicate more important components.}
    \label{fig:summary-results}
\end{figure*}

Figure~\ref{fig:summary-results}(a) summarizes the main comparison. VLSR achieves the best overall performance among image-input methods and outperforms the strongest symbolic baseline. Relative to Qwen3.5-4B + MSR-SFT, accuracy rises from 0.767 to 0.842, F1 from 0.679 to 0.786, and balanced accuracy from 0.745 to 0.829. Regression also improves. These results establish overall performance; the controlled tests below ask whether learning where to look explains the gain.

Increasing model scale or generating longer rationales is not sufficient. Qwen3.5-27B improves over smaller untuned visual baselines on classification but remains behind task-specific models, and Think-style reasoning does not consistently improve regression. VLSR instead learns where the relevant chemistry lies before applying latent reasoning.

\subsection{Efficiency Analysis}

Figure~\ref{fig:summary-results}(b) compares inference throughput on the full test set. VLSR keeps intermediate reasoning within a fixed-depth latent workspace and decodes only the final answer. It processes 39.84 samples/s, compared with 4.13 samples/s for image-input Qwen3.5-4B-SFT with textual reasoning, corresponding to a $9.6\times$ throughput improvement. Thus, the latent workspace improves efficiency without removing the learned localization that supports prediction.

\subsection{Visual Analysis}
\label{sec:visual-analysis}

\begin{figure}[t]
    \centering
    \includegraphics[width=\columnwidth]{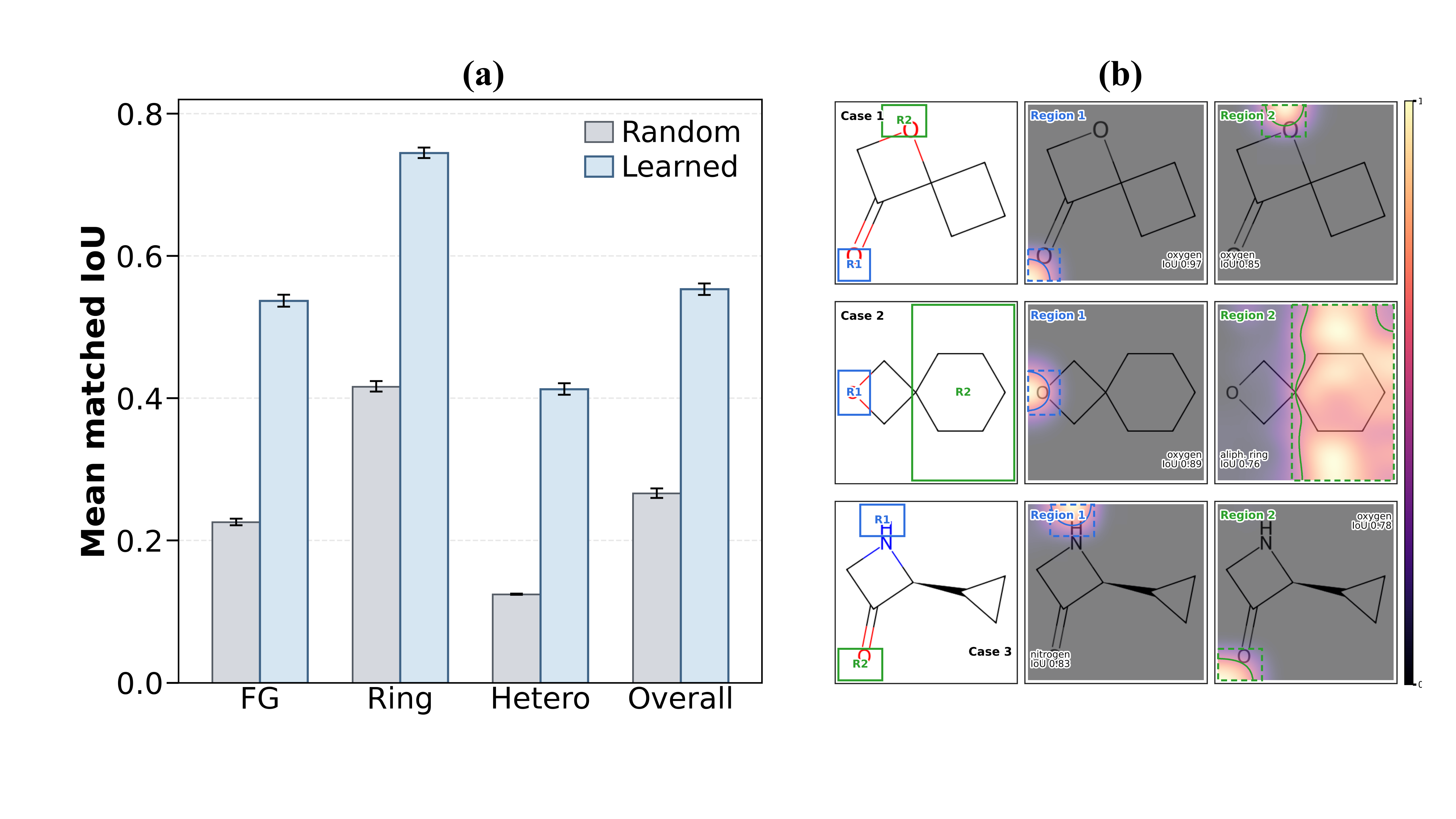}
    \caption{
    Chemical-region localization.
    (a) Matched IoU with held-out region annotations.
    (b) Different queries localize distinct structures within the same
    molecule.
    }
    \label{fig:localization-analysis}
\end{figure}

\paragraph{Localization.}
Matched IoU measures whether each learned query overlaps an annotated chemical
region. On held-out molecules, learned queries outperform random regions for
functional groups, rings, and hetero-atom regions
(Figure~\ref{fig:localization-analysis}(a)). Distinct queries also attend to
different annotated structures within the same molecule
(Figure~\ref{fig:localization-analysis}(b)). These results indicate that the
queries recover localized chemical structures rather than only encoding the
global molecular image.

\paragraph{Occlusion intervention.}
We further test whether the localized regions contain prediction-relevant
information by masking either the attention-ranked proposal or a
size-matched random region at inference time. All intervention conditions use
the same samples, checkpoint, prompts, and decoding configuration.
Learned-region masking reduces accuracy to 0.789 and raw-pooled Pearson
correlation to 0.576, compared with
$0.791$ and $0.626$, respectively, for random masks averaged over 20
deterministic draws using seeds 17--36
(Table~\ref{tab:visual-robustness}).
The accuracy gap is negligible, but the larger Pearson drop under learned-region
masking indicates that the proposals carry more regression-relevant evidence
than equally sized random regions. Occlusion alone does not establish a causal
chemical mechanism.

\paragraph{Relational reasoning.}
To separate localization from reasoning over localized evidence, we replace
learned image regions with RDKit-derived region descriptors and their relative
spatial relationships. Adding these descriptors to SMILES increases Pearson
correlation from 0.639 to 0.674. Removing chemical regions yields a Pearson
correlation of 0.600, while random region inputs yield 0.604. Combining
SMILES with image features achieves 0.793 accuracy and 0.674 Pearson
correlation. None of these controls
matches VLSR, which achieves 0.842 accuracy and 0.718 Pearson correlation.
Together with the localization and occlusion results, these controls support
the proposed pipeline: VLSR identifies chemically organized regions and
relates them to the queried property within its latent workspace.

\paragraph{Renderer robustness.}
Finally, we test whether the learned pipeline depends on canonical RDKit
layouts. VLSR obtains 0.788 accuracy and 0.632 Pearson correlation on PyMOL2D
depictions, and 0.799 accuracy and 0.640 Pearson correlation under
3D-projected RDKit depictions, compared with 0.842 and 0.718 on canonical
images. The 3D projection introduces unusual orientations, compressed bonds,
crossings, and partial occlusion. The retained performance indicates that VLSR
can recover useful local structure under substantial depiction shifts.

\begin{table}[t]
\centering
\small

\begin{tabularx}{\columnwidth}
{@{}X@{\hspace{3pt}}c@{\hspace{3pt}}c@{\hspace{3pt}}c@{\hspace{3pt}}c@{}}
\toprule
Setting
    & Acc $\uparrow$
    & F1 $\uparrow$
    & Bal. Acc $\uparrow$
    & Pearson $\uparrow$ \\
\midrule

\multicolumn{5}{@{}l}
{\textit{Input representation controls}} \\

SMILES only
    & 0.773
    & 0.702
    & 0.762
    & 0.639 \\

SMILES + Chem desc
    & 0.777
    & 0.705
    & 0.765
    & 0.674 \\

Remove chemical regions
    & 0.770
    & 0.703
    & 0.742
    & 0.600 \\

Random region input
    & 0.763
    & 0.706
    & 0.732
    & 0.604 \\

SMILES + Image
    & 0.793
    & 0.726
    & 0.781
    & 0.674 \\

\midrule

\multicolumn{5}{@{}l}
{\textit{VLSR under renderer shifts}} \\

VLSR (3D-proj RDKit)
    & 0.799
    & 0.721
    & 0.782
    & 0.640 \\

VLSR (PyMOL2D)
    & 0.788
    & 0.704
    & 0.769
    & 0.632 \\

\midrule

\multicolumn{5}{@{}l}
{\textit{Inference-time localization occlusion}} \\

Random mask
    & 0.791
    & 0.720
    & 0.776
    & 0.626\\

Learned mask
    & 0.789
    & 0.715
    & 0.773
    & 0.576 \\
    
VLSR 
    & \textbf{0.842}
    & \textbf{0.786}
    & \textbf{0.829}
    & \textbf{0.718} \\
\bottomrule
\end{tabularx}

\caption{Input controls and renderer robustness. RDKit descriptors bypass
localization, whereas patch controls lack explicit region grouping. VLSR learns
localized regions and spatial relations for joint reasoning. Renderer shifts
alter molecular layouts, with 3D projection adding overlap and occlusion.}
\label{tab:visual-robustness}

\end{table}

\subsection{Zero-Shot Evaluation on Protein-Conditioned FEP-Derived Ligand Comparisons}

To evaluate whether VLSR can transfer molecular edit reasoning beyond the training distribution, we conduct a zero-shot evaluation on protein-conditioned ligand comparison tasks derived from FEP benchmarks. We combine 317 ligand-pair comparisons from the Schrödinger JACS and Merck subsets with 467 comparisons from additional Schrödinger pairwise datasets \citep{wang2015fep,hahn2021openffbenchmark,hahn2022benchmarkbestpractices,ross2023fep}, covering 28 protein targets and 784 ligand comparisons in total.

No FEP labels, affinity values, or protein contexts are used during training. Furthermore, no fine-tuning, calibration, or model selection is performed on this evaluation set. The model receives a ligand pair together with the corresponding target protein sequence as contextual information, but does not access binding pockets, ligand poses, docking structures, or three-dimensional protein--ligand complexes. To reduce overlap with training data, both ligands in every retained pair have maximum ECFP4 Tanimoto similarity below 0.7 with molecules used during CPT and SFT \citep{rogers2010ecfp}.

The task is formulated as directional ligand comparison: given two ligands and a target protein context, the model predicts which ligand has stronger binding affinity. Free-energy calculations are used only to derive pairwise labels and are never provided as supervision targets. Therefore, this experiment evaluates the transferability of depiction-based molecular edit reasoning under protein-conditioned contexts, rather than quantitative FEP prediction.

\begin{table}[t]
\centering
\small
\begin{tabular*}{\columnwidth}
{@{\extracolsep{\fill}}llc@{}}
\toprule
Model & Protocol & Acc $\uparrow$ \\
\midrule
\multicolumn{3}{@{}l}{SMILES-based Input} \\
DeepSeek-V4-Flash
    & Instruct & 0.540 \\
DeepSeek-V4-Flash + MSR
    & Instruct & 0.589 \\
DeepSeek-V4-Flash
    & Think & 0.533 \\
DeepSeek-V4-Flash + MSR
    & Think & 0.520 \\
Qwen3.5-4B + MSR-SFT
    & Instruct & 0.628 \\
LatentChem
    & Latent & 0.432 \\
\midrule
\multicolumn{3}{@{}l}{Image-based Input} \\
ChemDFM-X
    & Instruct & 0.363 \\
ChemVLM-26B
    & Instruct & 0.377 \\
VLSR (Ours)
    & Implicit Latent & \textbf{0.691} \\
\bottomrule
\end{tabular*}
\caption{
Accuracy on the zero-shot protein-conditioned ligand comparison benchmark.
The task evaluates directional affinity ranking from ligand pairs and target protein context.
No FEP labels or affinity values are used during training.
}
\label{tab:independent-set}
\end{table}

As shown in Table~\ref{tab:independent-set}, VLSR achieves 0.691 accuracy, outperforming the strongest SMILES-based baseline, Qwen3.5-4B + MSR-SFT (0.628), by 6.3 percentage points. This result suggests that the learned localization and latent reasoning mechanism can generalize to ligand comparison scenarios where the property context differs from the training benchmark.

We further compare with structure-augmented reasoning baselines. In a tool-assisted setting, DeepSeek-V4-Flash combined with MSR provides textualized molecular structure descriptions. MSR improves the Instruct setting from 0.540 to 0.589, but does not consistently improve Think-style prompting. These results indicate that explicit structural descriptions can provide useful information, while textualized structure alone does not replace identifying and aligning molecular edits directly from visual representations.

This evaluation has several limitations. The task only measures pairwise affinity direction conditioned on protein sequence and does not model the physical determinants of protein--ligand interactions. Since no binding geometry, pocket information, or three-dimensional complex structure is provided, the result should not be interpreted as recovering protein--ligand physics or performing quantitative free-energy estimation. Instead, it measures whether molecular edit reasoning learned from two-dimensional depictions can transfer to a more challenging protein-conditioned comparison setting.

\begin{figure}[!t]
    \centering
    \begin{tabular}{@{}c@{\hspace{0.02\columnwidth}}c@{}}
        \includegraphics[width=0.47\columnwidth]{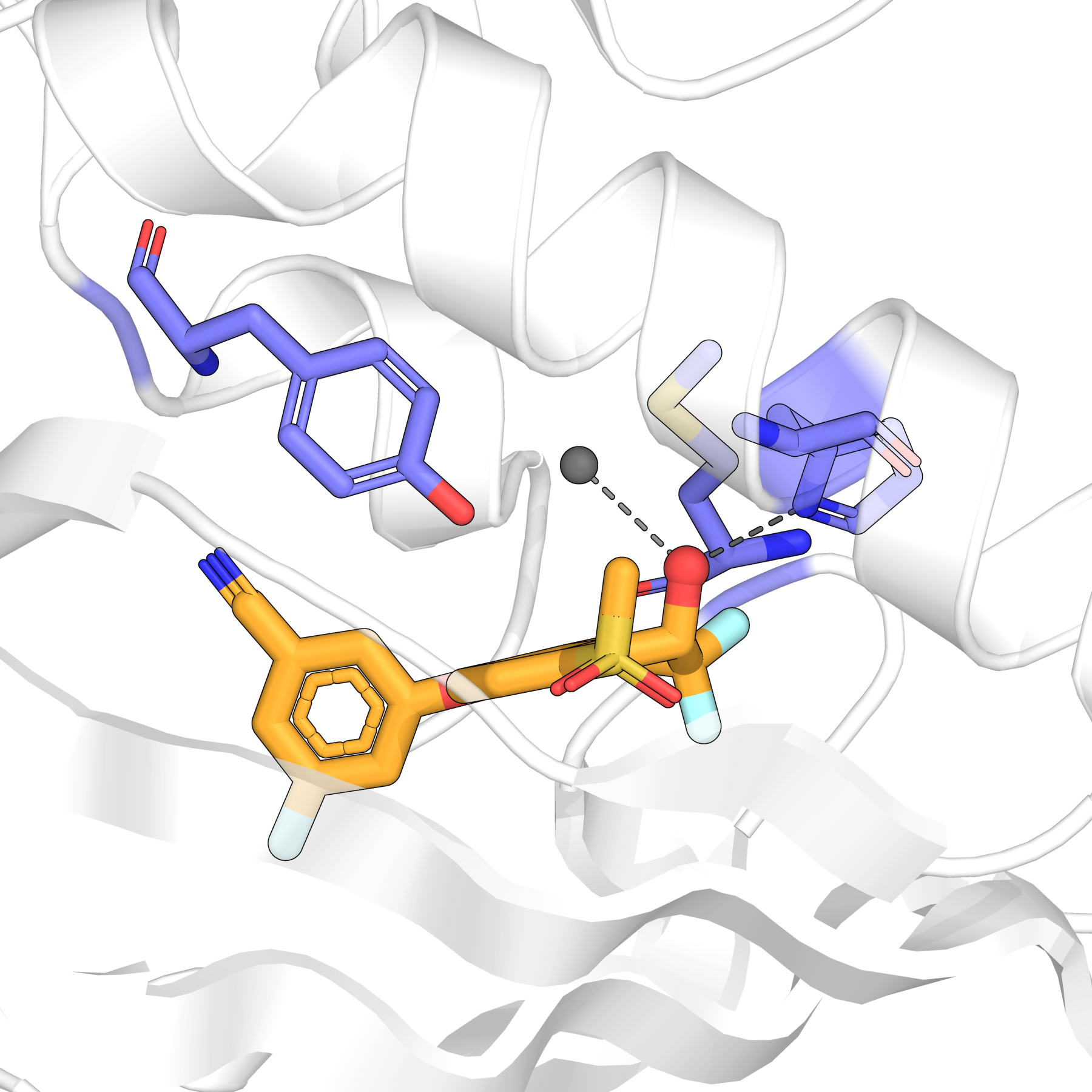} &
        \includegraphics[width=0.47\columnwidth]{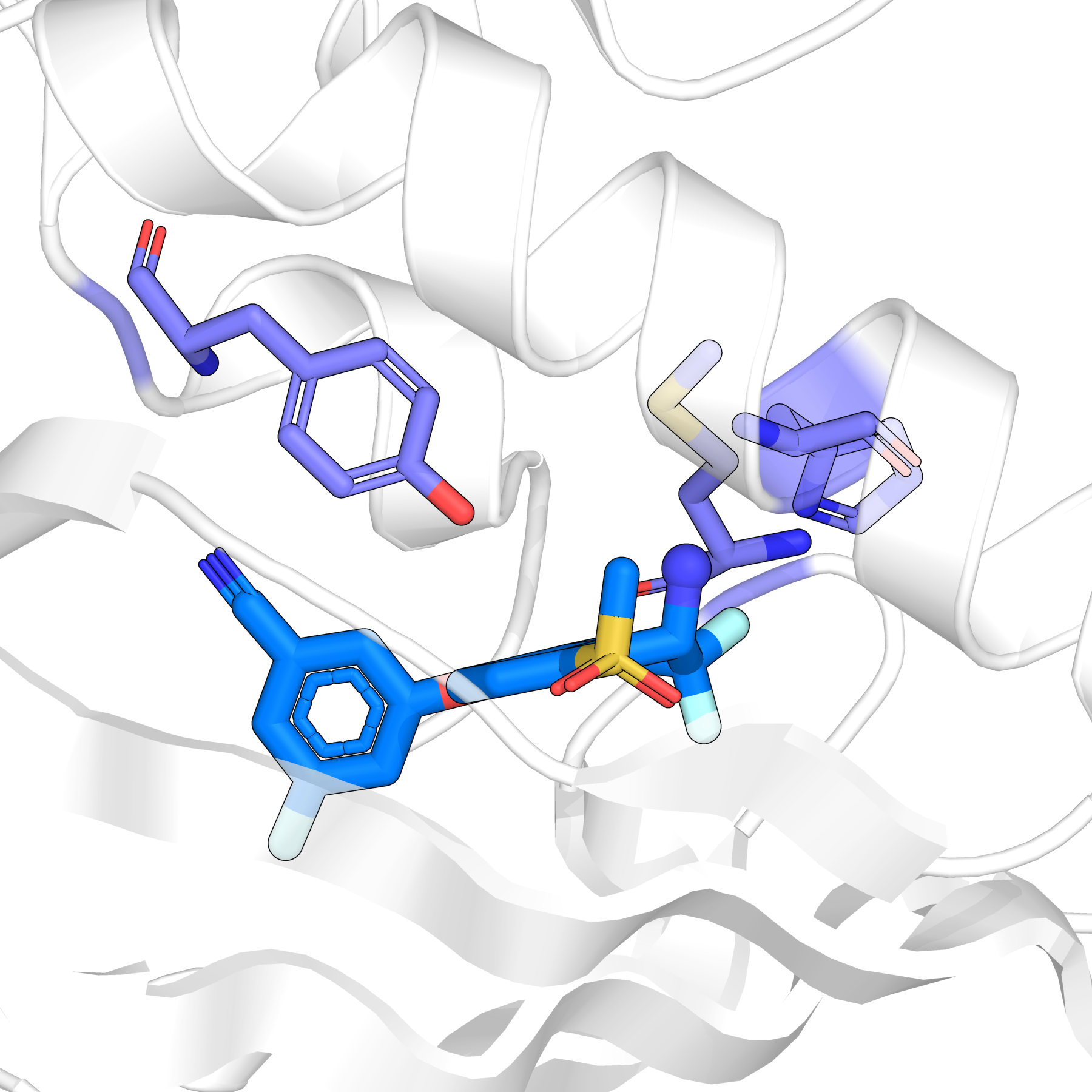} \\
        {\small (a) lig\_163 crystal pose} &
        {\small (b) lig\_165 pose projection}
    \end{tabular}

    \caption{Post hoc structural interpretation of a correctly
    predicted HIF-2$\alpha$ comparison. The lig\_163 crystal pose
    shows the edited site near His293 and a structural water;
    lig\_165 projects the OH$\rightarrow$NH$_2$ edit onto the same
    pose. The structure was not provided to VLSR.}
    \label{fig:hif2a-case}
\end{figure}

\paragraph{HIF-2$\alpha$ case study.}
VLSR correctly predicts the approximately 20-fold potency loss for the
lig\_163$\rightarrow$lig\_165 OH$\rightarrow$NH$_2$ edit
\citep{schindler2020fep}.
Mapping the pair to the PT2385-bound structure (PDB 5TBM)
\citep{wallace2016hif2a} places the
edited group near His293 and a structural water
(Figure~\ref{fig:hif2a-case}).
The lig\_165 pose is projected for interpretation and was not provided
to the model.

\subsection{Ablation Study}

Figure~\ref{fig:summary-results}(c) tests whether the learned regions contribute to the gain. Removing the learned regions causes a clear drop, and replacing them with random regions degrades performance further. Additional token capacity alone is therefore insufficient; the model benefits from finding useful local evidence.

Removing the workspace reduces classification and regression performance, showing that localization is not sufficient unless the model relates what it finds to the query. Removing edit alignment also hurts, particularly when corresponding structures appear at different image positions. Together, the ablations support the intended order: locate the evidence, establish correspondence when needed, and reason about its property effect. 

\subsection{Limitations}
VLSR uses RDKit-derived local chemical region boxes and labels as training
targets rather than inference inputs; it therefore does not eliminate
chemical supervision, and the annotation inventory may bias the learned
regions. CPT is used only for molecular-image understanding and contains no
property-reasoning supervision. Completely excluding molecular overlap with
general chemical pretraining corpora is impractical for foundation models;
we therefore evaluate task generalization using scaffold-controlled splits
and held-out molecules. Finally, 2D depictions cannot recover conformational
energetics or protein--ligand contacts absent from the input. The FEP-derived
results thus demonstrate directional transfer rather than full recovery of
these physical effects. Because region supervision is defined by rectangular
boxes, overlapping motifs and diffuse electronic effects may not correspond to
a single target. Matched IoU should therefore be read as spatial agreement,
while chemically valid counterfactual edits and multi-renderer consistency
offer important future tests of causal faithfulness. This limitation is most
relevant when a queried property depends on distributed electronic context
rather than a compact functional group. Such cases may require larger or
relationally defined regions than the present annotation inventory provides.

\section{Conclusion}
\label{sec:conclusion}

VLSR establishes a localize-then-reason framework for molecular visual
reasoning, learning chemically relevant regions before integrating their
property effects within a compact latent workspace. Across scaffold-controlled
evaluations, this design improves both classification and regression
performance, while the ablation results show that region localization, edit
alignment, relative spatial relationships, and latent reasoning make complementary
contributions. By avoiding lengthy textual rationales and decoding only the
final answer, VLSR achieves \(9.6\times\) higher inference throughput than the
corresponding image-input reasoning baseline. Its robustness across molecular
depiction styles and its transfer to protein-conditioned comparisons without
additional training further indicate that the learned regional representations
capture reusable chemical evidence. Overall, these results demonstrate that
explicitly connecting visual localization with property reasoning provides an
accurate, efficient, and inspectable foundation for molecular prediction from
2D depictions.

\section*{Acknowledgments}

This work was supported by funds from the National Institutes of Health
(R01GM147673, R01GM149705) and the National Science Foundation (1955260). The
authors would like to thank the computing resources provided by the Center for
Research Computing (facility RRID: SCR 022735) at the University of Pittsburgh
(NSF award number OAC-2117681), and the Pittsburgh Supercomputer Center (grant
number BIO210185).

\bibliography{vlsr_references}


\end{document}